\documentclass[journal]{IEEEtran}
\usepackage{xcolor}
\usepackage{cite}
\usepackage{hyperref}
\usepackage{amsmath}
\usepackage{multirow}
\usepackage{array}
\usepackage{moreverb,url}

\usepackage{algorithm,algorithmic}%

\usepackage[pdftex]{graphicx}
\ifCLASSINFOpdf
\else
\fi
\usepackage{amsfonts}

\usepackage{stfloats}
\begin{document}
%
\title{A Dynamic Spatial-temporal Attention Network for Early Anticipation of Traffic Accidents}

%
%
%

\author{Muhammad Monjurul Karim,
        Yu Li,
        Ruwen Qin\IEEEauthorrefmark{1}, {\it Member},
        Zhaozheng Yin, {\it Senior Member}
\thanks{Muhammad Monjurul Karim, Yu Li, and Ruwen Qin are with the Department
of Civil Engineering,  Stony  Brook  University, Stony Brook, NY 11794, USA.}
\thanks{Zhaozheng Yin is with the Department of Biomedical Informatics, Department of Computer Science, and AI Institute, Stony Brook University, Stony Brook, NY 11794, USA.}
\thanks{\IEEEauthorrefmark{1} Corresponding author: Ruwen Qin, email: ruwen.qin@stonybrook.edu}
\thanks{Manuscript received MM DD, YYYY; revised MM DD, YYYY.}}

%
%

\markboth{Journal of \LaTeX\ Class Files,~Vol.~XX, No.~X, MM~YYYY}%
{Shell \MakeLowercase{\textit{et al.}}: Bare Demo of IEEEtran.cls for IEEE Journals}
%



\maketitle

\begin{abstract}
The rapid advancement of sensor technologies and artificial intelligence are creating new opportunities for traffic safety enhancement. \textcolor{blue}{Dashboard cameras (dashcams) have been widely deployed on both human driving vehicles and automated driving vehicles. A computational intelligence model that can accurately and promptly predict accidents from the dashcam video will enhance the preparedness for accident prevention.} The spatial-temporal interaction of traffic agents is complex. Visual cues for predicting a future accident are embedded deeply in dashcam video data. Therefore, the early anticipation of traffic accidents remains a challenge. Inspired by the attention behavior of humans in visually perceiving accident risks, this paper proposes a Dynamic Spatial-Temporal Attention (DSTA) network for the early accident anticipation from dashcam videos. The DSTA-network learns to select discriminative temporal segments of a video sequence with a Dynamic Temporal Attention (DTA) module. It also learns to focus on the informative spatial regions of frames with a Dynamic Spatial Attention (DSA) module. A Gated Recurrent Unit (GRU) is trained jointly with the attention modules to predict the probability of a future accident.  The evaluation of the DSTA-network on two benchmark datasets confirms that it has exceeded the state-of-the-art performance. A thorough ablation study that assesses the DSTA-network at the component level reveals how the network achieves such performance. Furthermore, this paper proposes a method to fuse the prediction scores from two complementary models and verifies its effectiveness in further boosting the performance of early accident anticipation. 
\end{abstract}

\begin{IEEEkeywords}
Dynamic temporal attention, dynamic spatial attention, model fusion, autonomous vehicle, human-inspired AI
\end{IEEEkeywords}

%
\IEEEpeerreviewmaketitle

\section{Introduction}
%
%
%
%

\IEEEPARstart{A}{utonomous} driving has drawn increasing attention and made significant achievements in recent years  \cite{eskandarian2019research, muhammad2020deep}. While autonomous driving provides convenience to people and addresses emerging needs from industry, it raises concerns about traffic accidents. \textcolor{blue}{From 2014 to September 1, 2021, 337 autonomous vehicle collisions were reported in California, USA \cite{ca_dmv_2021}}. Moreover, according to the 2018 global status report on road safety from World Health Organization, about 1.35 million people are killed in traffic accidents every year \cite{world2018global}. \textcolor{blue}{Accident anticipation is a desired safety-enhancement capability for not only autonomous vehicles but also countless human driving vehicles. Successful anticipation of accidents from the widely deployed dashcams even just a few seconds ahead would effectively increase the situational awareness of human drivers, Advanced Driver Assistance Systems (ADAS), and autonomous vehicles to trigger a higher level of preparedness for accident prevention}.

\textcolor{blue}{
Many causal factors contribute to traffic accidents \cite{li2021crash}, including but not limited to human factors, environmental conditions, road and traffic characteristics, temporal-spatial factors, and vehicle types \cite{singh2015critical}. Detecting accident causal factors can help improve the awareness of accident risks and develop methods to mitigate their negative impacts on safety. Dashcams capture many of these causal factors in the form of video data. Computational methods that can translate the easily obtained dashcam video data into the perception of accident risks are largely desired \cite{janai2020computer}.}

Some pioneering studies of computer vision-based accident anticipation use conventional recurrent neural networks to capture causal factors by distributing soft attention to agents in the traffic scene \cite{chan2016anticipating,zeng2017agent, suzuki2018anticipating,fatima2021global}. Dashcam videos contain not only relevant information but irrelevant information. Without explicitly considering the spatial and temporal importance of traffic agents and the driving context along with their dynamic changes, it is impossible to effectively learn the video representation for  anticipating accidents. 

\textcolor{blue}{
Humans can turn their attention to risky regions in the driving scene. They use their peripheral vision to assess the scene and then fixate on regions that seem of high salient values. Drivers’ choice of attended regions and their levels of attention to them (e.g., measured by the fixation duration) vary over time. Their attention level increases when they perceive accident risks visually. Inspired by humans’ dynamic visual attention in perceiving traffic risks, this paper developed a computer vision-based deep neural network for the early anticipation of traffic accidents. The network has embedded spatial-temporal attention modules that reinforce its ability to anticipate traffic accidents. The technical contributions of this paper are threefold:}
\begin{itemize}
    \item \textcolor{blue}{A new framework named Dynamic Spatial-Temporal Attention (DSTA) network that learns to dynamically attend to the salient temporal segments and spatial regions of the driving scene video for the early accident anticipation. The network has outperformed the state-of-the-art;}
    \item \textcolor{blue}{A new method named score-level late-fusion that can be implemented on any set of complementary trained networks. The fusion method achieves higher prediction accuracy than the constituent networks;
    }
    \item \textcolor{blue}{A detailed analysis of the applicability and limitations of representative traffic accident datasets.}
\end{itemize}

The remainder of this paper summarizes the literature in Section II, delineates the proposed DSTA-network in Section III, and discusses the implementation detail and the experimental assessment of the network in Section IV. At the end, the paper summarizes findings and important future work.

\section{The Literature}

Accident anticipation, in general, falls into the category of problems that predict the probability of a future event. This problem is studied thoroughly to anticipate human actions. The computer vision-based approach to human action anticipation commonly uses appearance features as cues for prediction, including those at the object, activity, and context levels, respectively \cite{koppula2015anticipating,vondrick2016anticipating,minguez2018pedestrian,xu2019temporal,liang2019peeking}. Other cues that may further strengthen the prediction ability include the temporal relationship of sub-activities or activity-related entities, and the spatial relationship between humans, other entities, and the environment \cite{li2021crash}. Furthermore, the design of the loss function for training prediction neural networks is a mechanism for encouraging early predictions \cite{sadegh2017encouraging}. \textcolor{blue}{While these studies have built a strong methodological foundation for the probabilistic prediction of future events, they analyze video data captured by static surveillance cameras, not applicable to mobile cameras mounted on vehicles.} 

Various efforts are made to anticipate traffic accidents from dashcam videos. Yao et al. \cite{yao2019unsupervised} developed an unsupervised approach that utilizes the ego-vehicle motion information to monitor and predict future locations of traffic agents. \textcolor{blue}{Takimoto et al. \cite{takimoto2019predicting} incorporated physical location data with video data to predict the occurrence of traffic accidents. Closely related to \cite{ma2016learning, sadegh2017encouraging}, Suzuki et al. \cite{suzuki2018anticipating} proposed an adaptive loss function for promoting the early anticipation of accidents. The loss function assigns penalty coefficients according to the achieved mean time-to-accident during training.}

\textcolor{blue}{Recently, attention mechanisms are receiving growing interest. Like humans, a neural network with attention units can learn to selectively concentrate on most relevant features. Chan et al. \cite{chan2016anticipating} introduced the dynamic soft-attention to the traffic accident anticipation, which fuses the weighted sum of object-level features with the frame-level features of each video frame. The weights represent the attention levels on different objects.  Inspired by \cite{chan2016anticipating}, Zeng et al. \cite{zeng2017agent} proposed a soft-attention RNN that models the nonlinear interaction between traffic agents and locates risky regions where the agents may involve in a future accident. Fatima et al. \cite{fatima2021global} introduced a feature aggregation block that calculates the weights for aggregating object-level features to capture the inter-object interactions. The above-discussed studies mainly focus on learning attentions to spatially distributed agents that may be related to accidents. The temporal importance of appearance features are ignored either at the frame-level, the object-level, or both. From a different application domain, Cui et al. \cite{cui2021dynamic} integrated both a spatial attention module and a temporal attention module with a GRU \cite{chung2014empirical} to estimate the state-of-health of batteries. The model in \cite{cui2021dynamic} processes the feature vector measured at any time using a one-dimensional (1D) convolution layer, assigns ``spatial" weights to filter kernels, and learns a set of temporal weights that do not update over time. This approach is not applicable to the recurring task of accident anticipation from 2D video frames.}

Very recently, Bao et al. \cite{bao2020uncertainty} used a graph convolutional recurrent neural network (GCRNN) to capture the spatial-temporal relations among candidate objects. By measuring the spatial distance between objects in each frame, a graph is created to capture their spatial relationships. However, spatial distances between objects in a frame do not capture their true spatial relationships in the real world. The study further used a folded vector of all the hidden representations as temporal attention. Although it is beneficial to model training to a certain extent, temporal attention does not update dynamically to reflect the latest temporal information received.

\section{Methodology}

\begin{figure}[tb]
\centering
\includegraphics[width=0.95\columnwidth]{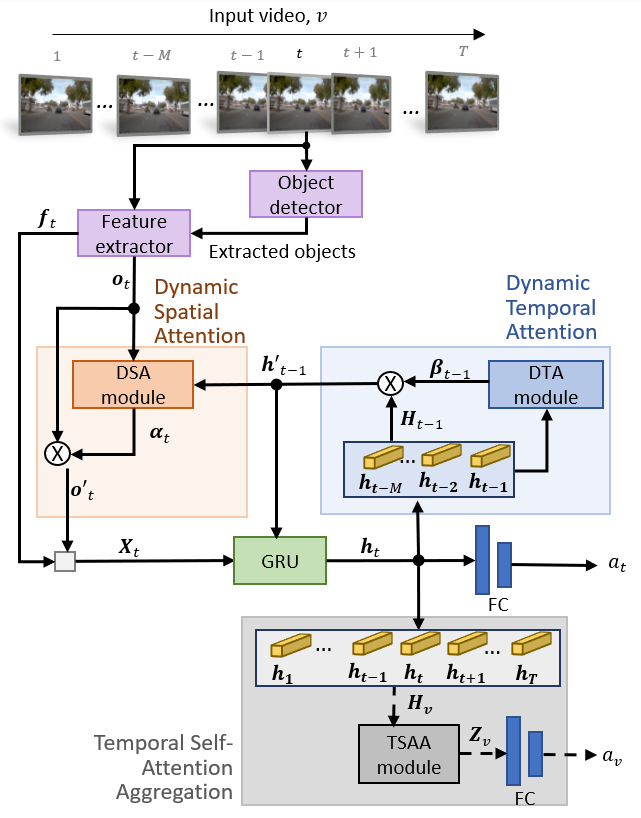}
\caption{Overview of the DSTA-framework}
\label{fig_overview}
\end{figure}

\textcolor{blue}{
The proposed network for accident anticipation integrates a Gated Recurrent Unit (GRU) with a Dynamic Spatial Attention (DSA) module, a Dynamic Temporal Attention (DTA) module, and Temporal Self-Attention Aggregation (TSAA) module, as Fig.\ref{fig_overview} illustrates. First, the network reads a dashcam video. Each frame of the video flows into an object detector to get multiple objects detected. Then, a feature extractor extracts both the frame-level features and object-level features. The weighted aggregation of the object-level features is concatenated with the frame-level features, becoming the overall feature input to the GRU. The GRU reads the input feature and the weighted aggregation of multiple hidden representations in the past to generate the hidden representation of the current frame. This hidden representation is used to predict the probability of seeing an accident in future frames. The DTA module learns the attention weights for aggregating hidden representations in the past, and the DSA module learns the attention weights for aggregating object-level features. Additionally, the appended auxiliary network TSAA learns the weights to aggregate all the hidden representations of each training video to predict the class of the video only in the training phase. Details of the proposed network are delineated below.}

\subsection{\textcolor{blue}{Dynamic Spatial-Temporal Attention Framework}}
\label{sub:dsta}

\textcolor{blue}{The Dynamic Spatial-Temporal Attention (DSTA) network uses a GRU, a simple yet well-performed type of recurrent neural network \cite{chung2014empirical}, to recurrently predict the probability of seeing an accident in future frames. The spatial-temporal relations of objects and the context information provide important cues for the accident anticipation. The GRU integrates a DSA module and a DTA module to learn the impact of the spatial-temporal relations to accident risks.}


\subsubsection{\textcolor{blue}{Feature Extraction and Aggregation}}

\textcolor{blue}{
An object detector detects spatially distributed objects from each video frame and keeps the top $N$ objects of the highest detection score for consideration. The VGG-16 \cite{Simonyan15} feature extractor extracts both frame-level and object-level features from each frame  through multiple convolution and pooling operations. Then, two fully connected layers flatten the extracted feature maps to become a $D$ dimensional feature vector. This study adds an additional fully connected layer to further reduce the dimension of the feature vector to $d$ ($<D$). Therefore, the extracted object-level features are $\boldsymbol{o}_t\in\mathbb{R}^{d\times N}$, and the frame-level features are $\boldsymbol{f}_t\in\mathbb{R}^{d}$. After passing through the DSA module to be introduced in the next section, $\boldsymbol{o}_t$ is turned into a weighted aggregation,  $\boldsymbol{o}^\prime_t \in\mathbb{R}^{d}$. Then, $\boldsymbol{o}^\prime_t$ is concatenated with the frame-level feature $\boldsymbol{f}_t$,
\begin{equation}
    \boldsymbol{X}_t = [\boldsymbol{o}'_t;\boldsymbol{f}_t],
\end{equation}
to form the overall feature vector of each frame, $\boldsymbol{X}_t\in\mathbb{R}^{2d}$.}

\subsubsection{Dynamic Spatial Attention (DSA)}
\label{sub:sa}

\textcolor{blue}{Attentions to the spatially distributed objects in a frame are unequal. This fact is modeled by the spatial attention weights, $\boldsymbol{\alpha}_t\in\mathbb{R}^{N}$, calculated using the weighted aggregation of hidden representations from the last step, $\boldsymbol{h}'_{t-1}$, to be introduced in the next section and the object-level features extracted from the current frame, $\boldsymbol{o}_{t}$: 
\begin{equation}
\boldsymbol{\alpha}_{t} =\gamma(\boldsymbol{W}_{sa}^\text{T}\tanh(\boldsymbol{W}_g \boldsymbol{h}^\prime_{t-1} +\boldsymbol{W}_\theta \boldsymbol{o}_{t} +\boldsymbol{B}_\theta)),
\label{eq:u_weight}
\end{equation}
where
$\boldsymbol{W}_{sa}\in\mathbb{R}^{d}$, $\boldsymbol{W}_g\in\mathbb{R}^{d\times d}$, $\boldsymbol{W}_\theta\in\mathbb{R}^{d\times d}$, and $\boldsymbol{B}_\theta\in\mathbb{R}^{d}$ are parameters of the DSA module, \textcolor{blue}{which are learned during training}. \textcolor{blue}{$\tanh$ is an activation function that regulates the values flowing through the network, and} $\gamma$ in this section is the softmax operator \textcolor{blue}{that normalizes the attention scores as spatial attention weights.}}
\textcolor{blue}{
Given the weights, the object-level features are turned into a weighted aggregation, $\boldsymbol{o}'_{t}\in\mathbb{R}^{d}$:
\begin{equation}
\boldsymbol{o}'_t =\boldsymbol{o}_{t}\boldsymbol{\alpha}_t.
\label{eq:attention_weight}
\end{equation}}

\subsubsection{Dynamic Temporal Attention (DTA)}

\textcolor{blue}{
Frames in a video are not equally important for the accident anticipation. Some frames may contain more discriminative information for accident prediction, whereas others mainly provide contextual information.} Motivated by this fact, the study designs the DTA module to provide the temporal attention weights for aggregating the hidden representations of the most recent $M$ frames.
Denote $\boldsymbol{H}_{t-1}\in\mathbb{R}^{d\times M}$ as the hidden representations of the past $M$ frames indexed by $t-M$ to $t-1$: 
\begin{equation}
    \boldsymbol{H}_{t-1}=[\boldsymbol{h}_{t-1}, \dots,  \boldsymbol{h}_{t-M}].
\end{equation}
The sliding window size $M$ is chosen experimentally. A very long temporal window may distract the network from the accident anticipation task and a very short window would not provide sufficient temporal context for the task. The temporal attention weights,  $\boldsymbol{\beta}_{t-1}\in\mathbb{R}^{d\times M}$, are computed as
\begin{equation}
\boldsymbol{\beta}_{t-1} = \gamma(\boldsymbol{W}_{ta} \tanh(\boldsymbol{H}_{t-1})),
\label{eq:temporal_attention}
\end{equation}
where $\boldsymbol{W}_{ta}\in\mathbb{R}^{d\times d}$ are learnable parameters. Then, $\boldsymbol{\beta}_{t-1}$ is used to turn $\boldsymbol{H}_{t-1}$ into a weighted aggregation, $\pmb{h}^\prime_{t-1}\in \mathbb{R}^{d}$:
\begin{equation}
\boldsymbol{h}^\prime_{t-1}=\langle\boldsymbol{\beta}_{t-1},\boldsymbol{H}_{t-1}\rangle_r,
\label{eq:temporal_energy}
\end{equation}
where $\langle\,,\,\rangle_r$ represents the row-wise inner product of two equally-sized matrix.

\subsubsection{\textcolor{blue}{Spatial-Temporal Relational Learning with GRU}}

\textcolor{blue}{At any frame $t$, the GRU takes the feature vector $\pmb{X}_t$ and the hidden representation $\pmb{h}'_{t-1}$ as the inputs to obtain the hidden representation of the current frame, $\pmb{h}_t$. GRU has two gates, a reset gate and an update gate, which retain the most relevant information, $\pmb{g}_t^{(r)}$ and $\pmb{g}_t^{(u)}$, respectively from the video sequence by filtering out irrelevant information. The data flow through the GRU are expressed mathematically in equations (\ref{eq:resetgate}-\ref{eq:ht}):
}
\begin{equation}  
\textcolor{blue}{\pmb{g}_t^{(r)} = \sigma(\pmb{W}_g^{(r)}\pmb{X}_t + \pmb{B}_g^{(r)}\pmb{h}'_{t-1}),}
\label{eq:resetgate}
\end{equation} 
\vspace{-1em}
\begin{equation} 
\textcolor{blue}{\pmb{r}_t =  \tanh (\pmb{W}_r \pmb{X}_t+\pmb{B}_r(\pmb{g}_t^{(r)} \circ \pmb{h}'_{t-1}) ), }
\end{equation}
\vspace{-1em}
\begin{equation} 
\textcolor{blue}{\pmb{g}_t^{(u)} = \sigma(\pmb{W}_g^{(u)}\pmb{X}_t + \pmb{B}_g^{(u)}\pmb{h}'_{t-1}),} 
\end{equation} 
\vspace{-1em}
\begin{equation} 
    \textcolor{blue}{\pmb{h}_t = (1- \pmb{g}_t^{(u)}) \circ \pmb{r}_t  + \pmb{g}_t^{(u)} \circ \pmb{h}'_{t-1}, }
\label{eq:ht}
\end{equation}
\textcolor{blue}{wherein $\sigma$ represents the sigmoid activation function, $\circ$ is the element-wise product operator, $\pmb{W}$'s and $\pmb{B}$'s ($\in \mathbb{R}^{d\times d}$) are learnable parameters.}

\textcolor{blue}{After passing two fully connected layers, $\phi$, the hidden representation $\pmb{h}_t$ is turned into the scores of 
positive and negative classes. The scores are further normalized by the softmax operator to become the accident anticipation probability, $a_t$:}
\begin{equation} 
    \textcolor{blue}{a_t = \gamma[\phi(\phi(\pmb{h}_t;\pmb{W}_0,\pmb{B}_0);\pmb{W}_1,\pmb{B}_1)]}, 
\label{eq:at}
\end{equation}
\textcolor{blue}{where $\pmb{W}$'s and $\pmb{B}$'s ($\in \mathbb{R}^{d\times d}$) are learnable parameters of the fully connected layers.}

\subsubsection{\textcolor{blue}{Temporal Self-Attention Aggregation (TSAA)}}
\label{sub:saa}

To help train the hidden layers of GRU better, the auxiliary module TSAA is included in the training stage only. The TSAA module provides learnable weights $\boldsymbol{W}_{saa}\in\mathbb{R}^{T}$ to perform a temporal self-attention aggregation \cite{vaswani2017attention} of all the $T$ hidden representations of each training video to predict the video class. \textcolor{blue}{As illustrated in Fig. \ref{fig_overview}, all the $T$ hidden representations of the training video indexed by $v$ are stored as a matrix $\pmb{H}_v$:}
\begin{equation}
    \textcolor{blue}{\boldsymbol{H}_v=[\boldsymbol{h}_{1},\dots,\boldsymbol{h}_{T}]_v,}
\end{equation}
\textcolor{blue}{Then, the self-attention operation is applied to obtain the weighted aggregation of all hidden representations, $\pmb{Z}_v\in\mathbb{R}^{d}$: }
\begin{equation}
    \textcolor{blue}{\boldsymbol{Z}_{v} = \boldsymbol{H}_v \gamma(\boldsymbol{H}^\text{T}_v \boldsymbol{H}_v) \boldsymbol{W}_{saa},}
\end{equation}
\textcolor{blue}{where $\boldsymbol{W}_{saa}\in \mathbb{R}^\text{T}$ are learnable parameters of TSAA.}
\textcolor{blue}{This aggregated video-level representation flows into two additional fully connected layers and one softmax operator to predict the video-level score, $a_v$:}
\begin{equation} 
    \textcolor{blue}{a_v = \gamma[\phi(\phi(\pmb{Z}_v;\pmb{W}_{v_0},\pmb{B}_{v_0});\pmb{W}_{v_1},\pmb{B}_{v_1})] ,
    \label{eq:av}}
\end{equation}
\textcolor{blue}{where $\pmb{W}_v$'s and $\pmb{B}_v$'s ($\in \mathbb{R}^{d\times d}$) are learnable parameters of the fully connected layers.}

\subsection{Training Procedure}
\textcolor{blue}{
The goal of the training process is to fit GRU, DSA, DTA, and TSAA to the training data by determining their learnable parameters described in Section \ref{sub:dsta}.} A video dataset of size $V$, indexed by $v$, is used to train the DSTA-network. Each video contains $T$ frames in total, captured at the rate of $f$ per second. The training dataset has two classes: positive and negative. \textcolor{blue}{If a video contains an accident starting from frame $\tau$ ($<T$), it belongs to the positive class and the video-level label is $l_v=1$. Otherwise, it belongs to the negative class and the label $l_v=0$ meaning that the video contains no accident. Prediction results of the DSTA-network are compared against the ground truth of training data to determine the loss that guides the learning process.}

$a_{t,v}$, calculated in Eq. (\ref{eq:at}),  denotes the probability that video $v$ contains an accident, anticipated by the DSTA-network at frame $t$ of the video. The frame-level loss function on the training dataset is calculated as:
\begin{equation}
\begin{aligned}
    \mathcal{L_F} =& \sum_{v=1}^V\left[-l_v\sum_{t=1}^Te^{-\max\left(\frac{\tau-t}{f},0\right)}\log(a_{t,v})\right.\\
    &\left. -(1-l_v)\sum_{t=1}^T\log(1-a_{t,v})\right].
\end{aligned}
\label{eq:frame_loss}
\end{equation}
The first term within the square bracket in Eq. (\ref{eq:frame_loss})  is the loss of a positive video, and the second term is the loss of a negative video. Negative videos use a regular binary cross entropy loss function. But the frame-level loss coefficient for positive videos increases exponentially as approaching the accident (e.g., $t<\tau$). After that, it reaches the same loss coefficient for negative videos. The use of the exponentially increasing loss coefficient for positive videos encourages the early anticipation of accidents.

The auxiliary TSAA module is trained to predict the probability that a video contains an accident using the video-level loss function:
\begin{equation}
\mathcal{L_V} =\sum_{v=1}^V \left[-l_v\log(a_v) -(1-l_v) \log(1-a_v)\right],
\label{eq:vid_loss}
\end{equation}
where $a_v$, calculated in Eq. (\ref{eq:av}), is the predicted probability that video $v$ is positive. The video-level loss function is a regular binary cross entropy loss function. The first term within the square bracket in Eq. (\ref{eq:av}) is the loss of a positive video and the second term is the loss of a negative video. 

The objective to achieve in training the DSTA-network is to minimize the following loss function:
\begin{equation}
\mathcal{L} = \mathcal{L_F} + w_a \mathcal{L_V}.
\label{eq:loss}
\end{equation}
Here, $w_a$ \textcolor{blue}{is a hyper-parameter that helps adjust the relative importance of the auxiliary loss $\mathcal{L_V}$ relative to the primary loss $\mathcal{L_F}$ to achieve an appropriate balance between them.} This study chooses 15 as the value for $w_a$ experimentally. \textcolor{blue}{Finally, the training progresses by backpropagating the loss $\mathcal{L}$ to update the learnable parameters of DSA, DTA, GRU, and TSAA.}

\subsection{Score-Level Late Fusion}
\label{subsec_fusion}

\textcolor{blue}{Any model for the early anticipation of traffic accidents, including the proposed DSTA-network, faces an unavoidable trade-off between the earliness of prediction and the correctness. Different models have their respective trade-off points. A deep ensemble of two or more complimentary models may further improve performance. Motivated by this hypothesis, the study proposes a method that fuses the prediction scores of two models that can complement each other to derive a more reliable prediction. This study found that models of the same architecture, but different parameters may complement each other, and so those of different architectures.}

Algorithm \ref{algorithm_fusion} below summarizes the proposed score-level late fusion method. Consider a video sequence flowing into two models. On each frame $t$, the two models give their own independent frame-level prediction scores, $a_t^{(1)}$ and $a_t^{(2)}$, respectively. If both models give scores greater than or equal to their respective classification thresholds, $\bar{a}^{(1)}$ and $\bar{a}^{(2)}$, the driving scene is anticipated to involve in a future accident confidently. Therefore, the fusion method takes the maximum score from the two models as the new anticipation probability. Similarly, if both the models return scores lower than their respective threshold values, the driving scene is anticipated to be safe confidently. Accordingly, the fusion method takes the minimum score from the two model as the new anticipation probability. If one model returns a score greater than or equal to its threshold while the other returns a score lower than its threshold, The two models give conflict classification. In such a case, the fusion method takes the mean of their scores as the new anticipation probability.

\begin{algorithm}
\caption{The Score-level Late Fusion Method}
\begin{algorithmic}
\STATE \textbf{Notation}: 
\STATE $T$: number of frames in a video sequence, indexed by $t$
\STATE $a_t^{(1)}$ and $a_t^{(2)}$: prediction scores by models 1 and 2, respectively, at frame $t$ 
\STATE $\bar{a}^{(1)}$ and $\bar{a}^{(2)}$: classification thresholds of models 1 and 2, respectively
\STATE\textbf{Input}: Frames of a video sequence
\STATE \textbf{Output}: Accident prediction scores, $\{a_t|t=1,\dots, T\}$
\vspace{0.03in}
\FOR{$t=1 \: to \: T$}
\IF {$a_t^{(1)} \geq \bar{a}^{(1)}$ \AND $a_t^{(2)} \geq \bar{a}^{(2)}$}
\STATE{$a_t = \max(a_t^{(1)},a_t^{(2)})$}
\ELSIF{\textcolor{blue}{{$a_t^{(1)} < \bar{a}^{(1)}$ \AND $a_t^{(2)} < \bar{a}^{(2)}$}}}
\STATE\textcolor{blue}{{$a_t = \min(a_t^{(1)},a_t^{(2)})$}}
\ELSE 
\STATE\textcolor{blue}{{$a_t = \text{mean}(a_t^{(1)},a_t^{(2)})$}}
\ENDIF
\ENDFOR
\end{algorithmic}
\label{algorithm_fusion}
\end{algorithm}

The threshold values of the two constituent models in Algorithm \ref{algorithm_fusion} are design parameters that impact the effectiveness of the fusion method. \textcolor{blue}{Different models have different prediction behavior. For example, one model may focus more on the earliness of prediction, while the other may focus more on the correctness. Therefore, using separate threshold values for the constituent models allow for taking advantage of their respective strengths. The values of $\bar{a}^{(1)}$ and $\bar{a}^{(2)}$ are determined in a numerical optimization approach. That is, given a pair of threshold values, the corresponding performance of the fusion method on the testing dataset is obtained. The decision is to select a pair of threshold values which leads to the best performance.}

\section{Implementation and Experimental results}
The study illustrates the implementation and evaluation of the proposed DSTA-network for the early accident anticipation on two publicly available datasets. 

\subsection{Datasets}


\subsubsection{Dashcam Accident Dataset (DAD) \cite{chan2016anticipating}} It contains a diverse set of videos captured across different cities in Taiwan. The frame rate of the video is 20 frames per second (fps). Each video lasts 5 seconds and thus consists of 100 ($T$) frames. This dataset has 1,130 negative videos without any accident and 620 positive videos that each contains an accident in the last 0.5 seconds. The training dataset includes 455 positive and 829 negative videos, whereas the testing dataset comprises 165 positive and 301 negative videos. 

\subsubsection{Car Crash Dataset (CCD) \cite{bao2020uncertainty}}
It is a dashcam dataset with diverse environmental attributes. \textcolor{blue}{It contains 4,500 videos. 80\% are training data and 20\% are testing data. For both training and testing data, the ratio of positive videos to negative videos is 1:2.} Each video lasts 5 seconds with 50 frames in total (i.e., fps is 10). For a positive video, the accident starting time is randomly placed in the last 2 seconds of the video.

\subsection{The Detail of Implementation}
The study implemented the proposed approach using PyTorch \cite{paszke2019pytorch}. Training and testing were performed using an Nvidia Tesla V100 GPU with 32GB of memory. In use of the DAD dataset, this study adopted the features of 19 ($N$) candidate objects and the frame-level appearance features provided by \cite{chan2016anticipating}. The candidate object classes are human, bicycle, motorbike, car, and bus which were detected using Faster R-CNN \cite{ren2015faster}. For the CCD dataset, the study also directly uses the features provided by \cite{bao2020uncertainty} for a fair result comparison, where candidate objects were detected using Cascade R-CNN \cite{cai2018cascade}. The dimension of VGG-16 features in both datasets is 4,096 ($D$). These features were passed through fully-connected embedding layers to reduce the dimension to 512 ($d$). The dimension of hidden representations returned by GRU is 512 too. \textcolor{blue}{Features and hidden representations of this dimension can represent input video frames well and they are learned within a reasonable amount of training time}. Parameters of the DSTA-network were initialized randomly, by taking values from a normal distribution with 0 mean and 0.01 standard deviation.
The temporal sliding window for the DTA module is 0.5 seconds ($M=0.5f$), which is verified experimentally to be suitable. 
\textcolor{blue}{Similar to Chan et al. \cite{chan2016anticipating}, a learning rate of 0.0001 and a batch size 10 were used to train the network. ReduceLROnPlateau was used as the learning rate scheduler. Adam optimizer was used to optimize the network for 60 epochs that are sufficiently long for this study.}

\subsection{Evaluation Metrics}
The evaluation of a model for the early accident anticipation needs to consider two aspects: the earliness and the correctness of prediction.

\subsubsection{Correctness}
The ability of a model to correctly anticipate accidents can be measured by its performance in classifying a set of testing videos. \textcolor{blue}{This study adopts the following three classic metrics \cite{goodfellow2016deep}.} 
\begin{itemize}
    \item Recall (R): the ratio of correctly predicted positives over the total number of positive videos
    \item Precision (P): the ratio of correctly predicted positive videos over the total number of positive predictions
    \item Average Precision (AP): The recall value and the precision value are changing when the classification threshold changes. The average precision is the area below the precision-recall curve: 
    \begin{equation}
        \text{AP}=\int \text{P}_{\text{R}}d\text{R},
    \end{equation}
\end{itemize}
\textcolor{blue}{where $\text{P}_{\text{R}}$ is the precision corresponding to the recall R. AP is independent of the choice of classification threshold.
}
\textcolor{blue}{A high recall value certainly is critical for accident anticipation due to the severe consequence of false negatives. But a very high recall could be unrealistic, especially when it is at the cost of very low precision. In accident anticipation studies \cite{chan2016anticipating,bao2020uncertainty}, 80\%  represents a reasonably good recall value for evaluating the corresponding precision.} This study adopts the precision corresponding to 80\% recall, denoted by P$_{80\text{R}}$, as another metric for the performance evaluation.

\subsubsection{Earliness}
The sooner a model can anticipate an accident, the more capable the accident prevention is. This study measures the earliness of accident anticipation using the following metrics{\cite{chan2016anticipating,bao2020uncertainty,zeng2017agent,suzuki2018anticipating}}:
\begin{itemize}
    \item Time-to-Accident (TTA) : The first time when the frame-level anticipation probability $a_t$ goes across the classification threshold $\bar{a}$ is the time to classify a video as a positive. TTA is the period between this time and the starting time of accident, $\tau$: 
    \begin{equation}
        \text{TTA}=\max\{\tau-t|a_t\ge \bar{a}, 1\le t\le \tau\}
    \end{equation}
    \item mean Time-to-Accident (mTTA): TTA is changing if the classification threshold $\bar{a}$ changes. The study thus calculates the mean value of TTA: 
    \begin{equation}
        \text{mTTA}=E[\text{TTA}].
    \end{equation}
\end{itemize}
mTTA is independent of the classification threshold. AP and mTTA are commonly used as a pair of metrics for the model assessment. TTA$_{80\text{R}}$ is the TTA at 80\% recall. TTA$_{80\text{R}}$ reflects the TTA at a relatively high requirement on recall.

\subsection{Evaluation of the DSTA-Network}
\label{sub:performance_evaluation}
\subsubsection{Experimental Evaluation}
\label{sub:exp_evaluation}
\begin{figure}[htbp]
\centering
\includegraphics[width=3.3in]{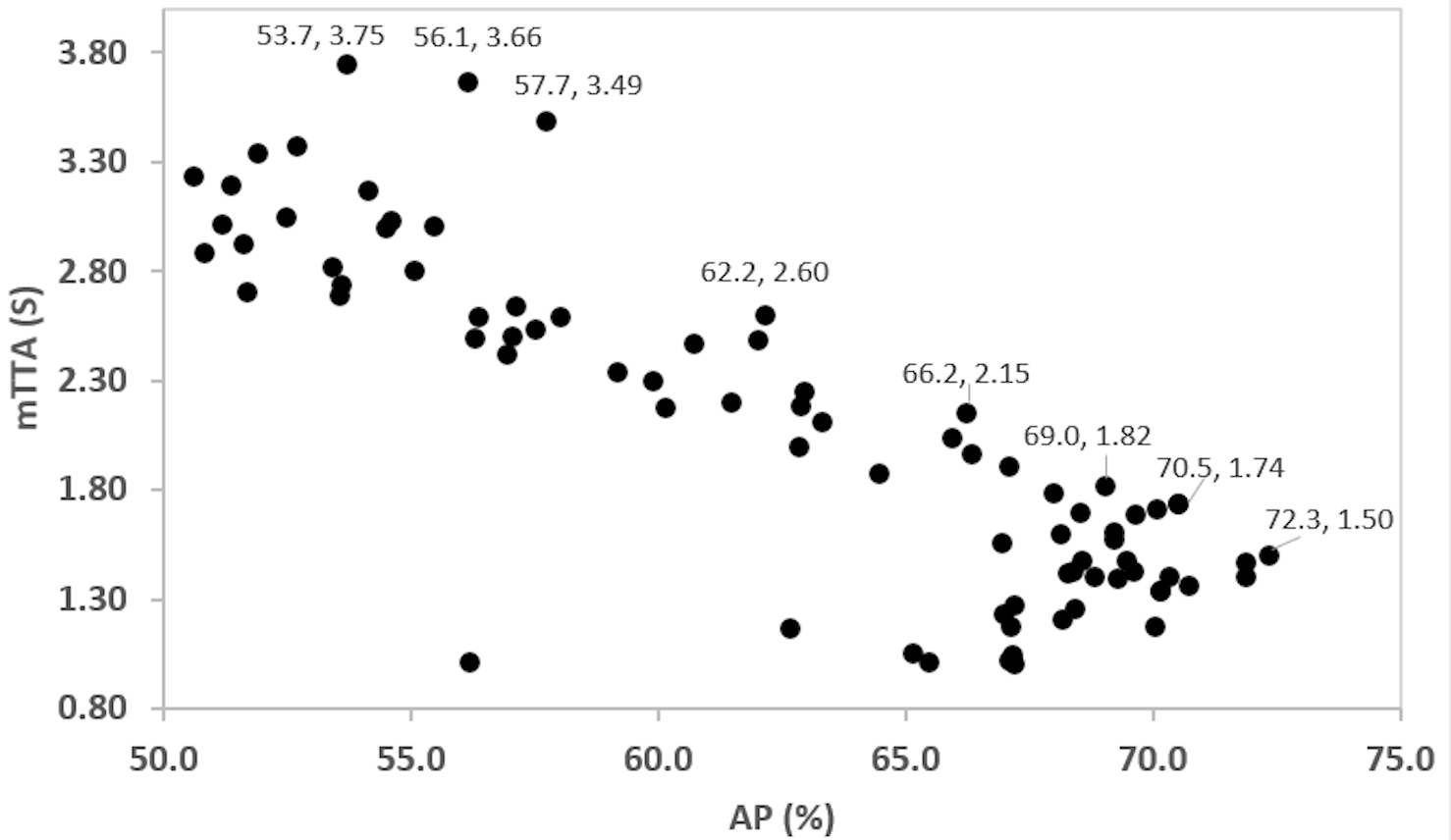}
\caption{Relationship between mTTA and AP on the DAD dataset}
\label{fig_mtta_ap}
\end{figure}

\textcolor{blue}{A learning behavior is observed from training the DSTA-network, which attempts to seek models that have larger AP and longer mTTA over multiple epochs. A trade-off between the earliness and the correctness of accident anticipation presents because the exponentially increasing loss coefficient for positive videos in Eq. (\ref{eq:frame_loss}) encourages predicting accidents earlier. The network learns a set of models that are positioned at different places on the mTTA-AP diagram. Models on the Pareto-optimal frontier \cite{censor1977pareto} are efficient models in that neither their mTTA nor AP can be further improved without sacrificing the other.} Fig. \ref{fig_mtta_ap} is the mTTA-AP diagram that displays a group of models the proposed DSTA-network learnt on the DAD dataset. It keeps models with at least 1 second of mTTA and at least 50\% of AP because a very short mTTA or a very low AP is not applicable in real-world applications. As seen from the figure, the efficient model with the longest mTTA (3.75 seconds) achieves 53.7\% AP.  Increasing the AP of an efficient model requires to shorten the mTTA. The rightmost point in this mTTA-AP diagram is the \textcolor{blue}{efficient model with the best AP}, which achieves 1.5 seconds of mTTA at 72.3\% AP. \textcolor{blue}{A similar relationship between mTTA and AP is also observed when training the DSTA-network on the CCD dataset.}

Selection of an efficient model from the Pareto-optimal frontier for the implementation should consider the specific need of users. Requirements on accident anticipation systems can vary, depending on the weather conditions, time of the day, and areas where vehicles are running \cite{li2021crash}. Some may demand super early prediction where false alarms are handled well. Others may require a high accuracy where earliness is not essential.

\subsubsection{Comparison to the State-of-the-Art Models}

The DSTA-network is compared to the state-of-the-art models targeting longer mTTA \cite{chan2016anticipating, zeng2017agent, suzuki2018anticipating, bao2020uncertainty}. TABLE \ref{tab:comparison} summarizes the results of the comparative study. The performance of \cite{chan2016anticipating, zeng2017agent, suzuki2018anticipating,bao2020uncertainty} are cited from \cite{zeng2017agent} and \cite{bao2020uncertainty}. \textcolor{blue}{On the DAD dataset, DSA published in 2016 \cite{chan2016anticipating} achieved 48.1\% AP and 1.34 seconds mTTA. Within four years, other studies have progressively increased the AP to 53.7\% and extended the mTTA to 3.66 seconds \cite{zeng2017agent, suzuki2018anticipating, bao2020uncertainty}. The proposed DSTA-network further leverages the AP by another 2.4\% and mTTA by 0.13 seconds, achieving 56.1\% AP and 3.66 secconds mTTA. This confirms that DSTA can improve both AP and mTTA on such a challenging dataset.} On the CCD dataset, the state-of-the-art performance seems has saturated already. But DSTA still outperforms it with a small margin, \textcolor{blue}{achieving 99.6\% AP with 4.87 seconds mTTA.} On both DAD and CCD datasets, the proposed DSTA-network has outperformed the existing state-of-the-art performance.

\begin{table}[htbp]
\renewcommand{\arraystretch}{1.3}
    \centering
    \caption{Comparison of models seeking longer mTTA on DAD and CCD}
    \begin{tabular}{>{\centering\arraybackslash}p{1.1cm}|r|>{\raggedright}p{2cm} |>{\raggedleft}p{1.2cm}|r}
        \hline
        Dataset &  \textcolor{blue}{Year} & Model & AP(\%) & mTTA(s)\\
        \hline
       \multirow{5}{*}{DAD}& \textcolor{blue}{2016}  & DSA \cite{chan2016anticipating}  & 48.1& 1.34 \\
       & \textcolor{blue}{2017} & L-RAI \cite{zeng2017agent}  & 51.4 & 3.01\\
       & \textcolor{blue}{2018} & adaLEA \cite{suzuki2018anticipating}  & 52.3 & 3.43\\
       & \textcolor{blue}{2020} &GCRNN \cite{bao2020uncertainty}  & 53.7 & 3.53\\
       \cline{2-5}
       & \textcolor{blue}{2021} &DSTA (Ours)  & \textbf{56.1} & \textbf{3.66}\\
       \hline
       \multirow{3}{*}{CCD} &\textcolor{blue}{2016}  & DSA \cite{chan2016anticipating}  & 99.6 & 4.52\\
       &\textcolor{blue}{2020} & GCRNN \cite{bao2020uncertainty} & 99.5 & 4.74 \\
       \cline{2-5}
       & \textcolor{blue}{2021} & DSTA (Ours) & \textbf{99.6} & \textbf{4.87}\\
       \hline
    \end{tabular}
    
    \label{tab:comparison}
\end{table}

TABLE \ref{tab:best_ap} further compares the best AP model that the DSTA-network attains to that of GCRNN \cite{bao2020uncertainty}. Compared to GCRNN, \textcolor{blue}{the best AP model of DSTA extends the mTTA by 13\% and TTA$_{80\text{R}}$ by 16\%. Meanwhile, it increases AP slightly, to 72.34 \%. Not to mention that extending the TTA even for a fraction of second can create better opportunities for accident prevention.} Results in TABLES \ref{tab:comparison} and \ref{tab:best_ap} verify the competitiveness of the DSTA-network in meeting various requirements for prediction correctness and earliness. \textcolor{blue}{The above-cited performances are associated with individual models. The proposed fusion method, to be discussed in Section \ref{sub:fusion_result}, will achieve further improvement.}

\begin{table}[htbp]
\renewcommand{\arraystretch}{1.3}
    \caption{Comparison of Best AP models on DAD}
    \label{tab:best_ap}
    \centering
    \begin{tabular}{>{\centering\arraybackslash}p{1.6cm}|>{\centering\arraybackslash}p{1.5cm}|>{\centering\arraybackslash}p{1.5cm}|>{\centering\arraybackslash}p{1.7cm}}
        \hline
         Model & AP(\%) & mTTA(s) & TTA$_{80\text{R}}$(s) \\
         \hline
         GCRNN \cite{bao2020uncertainty} & 72.22 & 1.33 & 1.56 \\
         DSTA (Ours) & \textbf{72.34} & \textbf{1.50} & \textbf{1.81} \\
         \hline
    \end{tabular}
    
\end{table}

\subsubsection{\textcolor{blue}{Ablation Study}}

An ablation study comprising 10 experiments is performed to assess the architecture of the proposed DSTA-network. A decrease in the model's performance due to the removal or replacement of a key component measures the component's contribution. TABLE \ref{tab:impact_attention} summarizes the best AP and corresponding mTTA attained in each of the experiments. 

\begin{table}[ht!]
\renewcommand{\arraystretch}{1.3}
    \caption{Ablation study on the DAD dataset}
    \label{tab:impact_attention}
    \centering
    \resizebox{\linewidth}{!}{%
    \begin{tabular}{c|l|c|c|c|r|r}
        \hline
         Experiment & RNN & DSA  & DTA & TSAA & AP (\%) & mTTA (s) \\
         \hline
        1 & GRU & \checkmark & \checkmark &\checkmark & 72.34 & 1.50  \\
         2 & GRU & \checkmark & \checkmark  ($M$=1) &\checkmark & 70.74 & 1.36 \\
         3 & GCRNN & \checkmark & \checkmark &\checkmark & 70.47 & 1.22 \\
         4 & LSTM & \checkmark   & \checkmark  & \checkmark  & 69.08 & 1.48  \\
         5 & GRU & \checkmark &\checkmark  & & 69.89 & 1.70 \\
         6 & GRU & \checkmark &  &\checkmark & 69.72 & 1.43 \\
         7 & GRU & &\checkmark  & \checkmark & 68.15 & 1.33 \\
         {8} & GRU &   &  &  \checkmark& {66.04} & {1.39}\\
         9 & GRU &    & \checkmark  &   & 65.99 & 1.42  \\
         10 & GRU & \checkmark  &  &  & 65.03 & 1.55 \\

         \hline
    \end{tabular}
    }
\end{table}

Experiment 1 is the DSTA-network with all the components in place. The attained AP is higher than that of any other experiment. It shall be mentioned that the DSTA-network can provide another model (see Fig. \ref{fig_mtta_ap}), whose AP is 70.5\% and mTTA is 1.74 seconds, longer than that of any other experiment in TABLE \ref{tab:impact_attention}. 

In experiment 2, the sliding window of the temporal attention model, $M$, is reduced to one frame (i.e., the last frame). This change reduces the AP for 1.6\%, which signifies the effectiveness of using several recent frames' hidden representations to learn the temporal relation. 

In experiment 3, a GCRNN replaces the GRU, resulting a drop of the AP for 1.87\% and a decrease of mTTA for 0.28 seconds. This is mostly because the edge weights of the GCRNN are less effective than the dynamic spatial attention weights that the DSTA-network learns. Similarly, the replacement of the GRU by an LSTM in experiment 4 lowers the AP for 3.26\% and reduces the mTTA slightly, for 0.02 seconds, \textcolor{blue}{despite of increasing the number of parameters.} Additionally, the training time of experiment 3 is about five times longer than the training times of experiments 1 and 4.

Experiment 5 drops the TSAA module, which decreases the AP for 2.45\% but increases the mTTA for 0.20 seconds. Experiment 6 drops the DTA module, resulting a decrease in the AP for 2.62\% and the mTTA for 0.07 seconds. Experiment 10 drops both modules, which decreases the AP for 7.31\% and but slightly increases the mTTA for 0.05 seconds. These experiments verify the importance of the temporal attention modules in improving AP. 


Compared to experiment 1, experiment 7 drops the DSA module. This change reduces the AP for 4.19\% and the mTTA for 0.17 seconds. Similar observations are seen from the comparison between experiments 9 and 5 and the comparison between experiments 8 and 6. These experiments verify that the DSA module in the DSTA-network is effective in increasing both the AP and the mTTA. 


The ablation study verifies the merits of the DSTA-network design. Each of the attention modules positively contributes to the improvement of AP. Their net effect on mTTA is also positive. \textcolor{blue}{This study also confirms that, given all these attention modules, GRU is a better choice than both LSTM and GCRNN.}

\subsubsection{Qualitative Results}
\begin{figure*}[htb]
\centering
\includegraphics[width=5.3in]{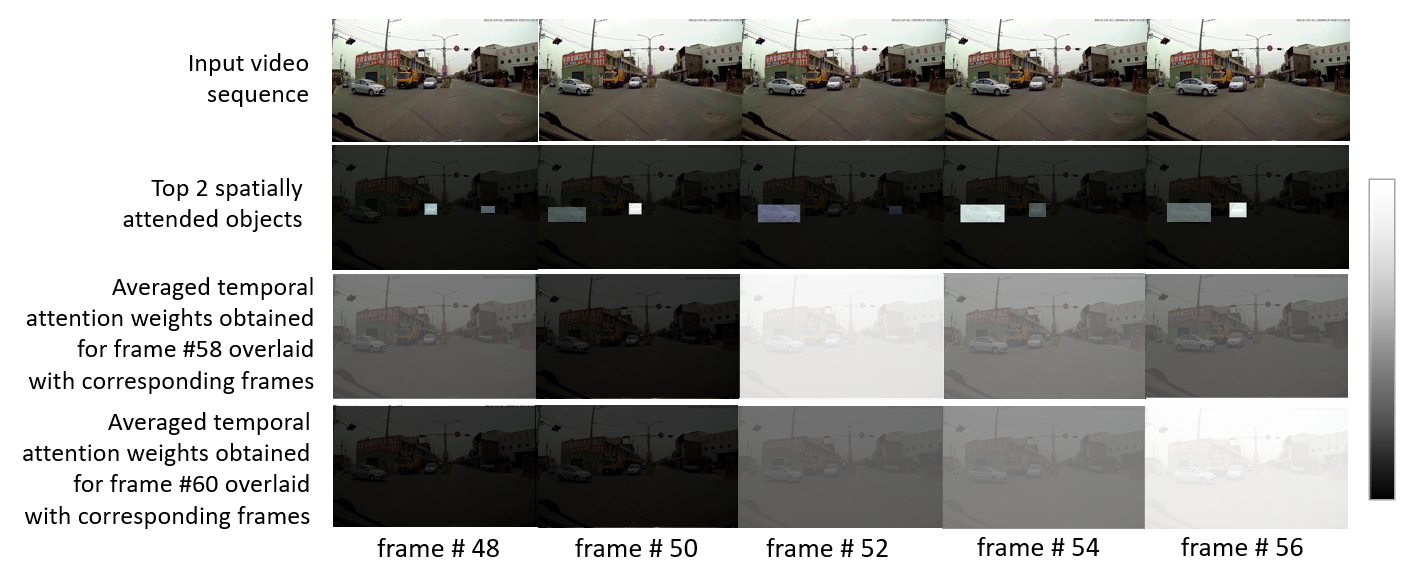}
\caption{Illustrating the dynamics of the spatial and temporal attentions of the DSTA-network}
\label{fig_TA}
\end{figure*}

The DSTA-network can focus on the most semantically significant spatial and temporal regions in a video sequence. Fig. \ref{fig_TA} illustrates the dynamics of the spatial and temporal attentions using a few sample frames from a positive video wherein two vehicles (cars) are approaching to each other in an intersection. The color bar on the right indicates the magnitude of the attention weights. 

The first row is the original sample frames. \textcolor{blue}{In the second row, the top two attended objects in each frame are indicated by colored rectangular boxes.} The object receiving the highest spatial attention weight and the magnitude of this weight may vary from one frame to another. \textcolor{blue}{In this example, the top two attended objects in frames \#50, \#54, and \#56 are traffic agents that will soon involve in an accident. In frames \#48 and \#52, only the object with the top attention weight is a related traffic agent. The observation indicates that the network will get more contextual information as time goes by and it will become more accurate in assigning the spatial attention.}
 
\textcolor{blue}{
According to Eq.(\ref{eq:temporal_attention}), the temporal attention assigned to the hidden representation of frame $t-j$, $\pmb{\beta}_{t-j}$, for $j=1,\dots,M$,  is a vector. This study calculates the mean value of this vector and uses it to visualize the temporal attention on this frame assigned at $t$.} The third row of Fig. \ref{fig_TA} shows the frames overlaid with their averaged temporal attention weights assigned at frame \#58. Frame \#52 receives the highest attention whereas frame \#50 receives the lowest attention. The fourth row further shows the frames overlaid with their averaged temporal attention weights obtained at frame \#60. The change of attention weights from row three to row four illustrates the dynamic nature of the temporal attention.


\begin{figure*}[htbp]
\centering
\includegraphics[width=5in]{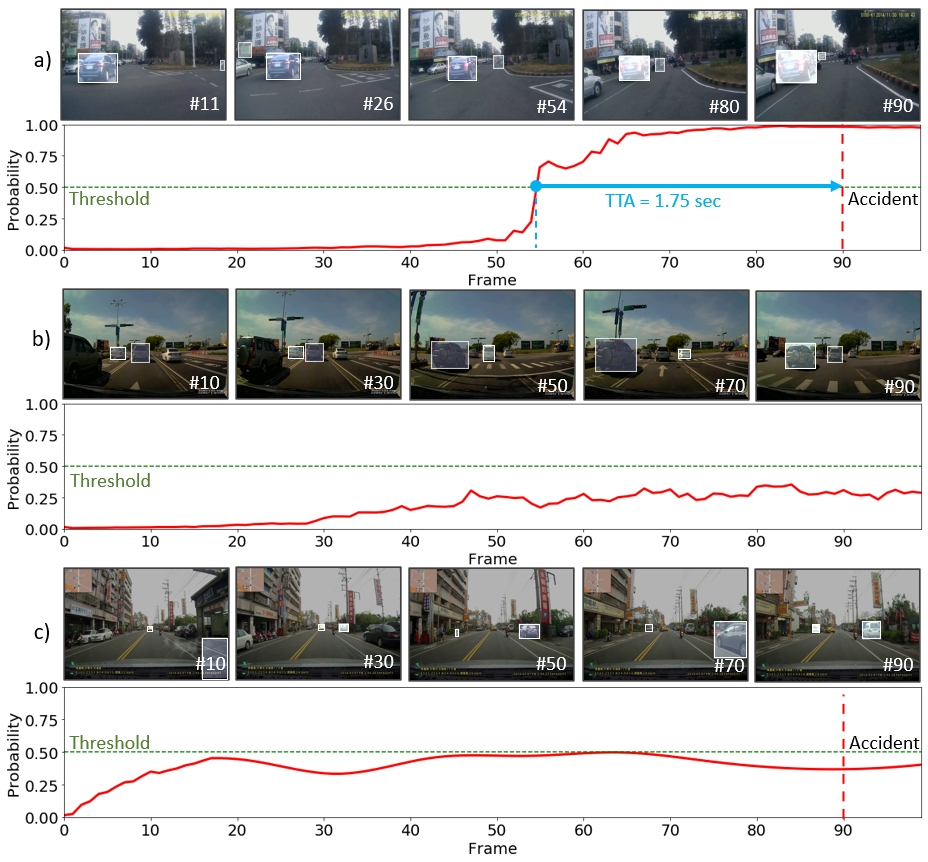}
\caption{\textcolor{blue}{Examples of accident anticipation on the DAD dataset: a) a true-positive sample, b) a true-negative sample, and c) a false-negative sample. For better visualization, white bounding boxes with a shade are added to the top two attended objects. The red curve indicates the prediction probability at each frame.}}
\label{fig_qualitative_1}
\end{figure*}

Fig. \ref{fig_qualitative_1} illustrates three examples of accident anticipation by the proposed DSTA-network. \textcolor{blue}{a) and b) are successful examples on analyzing a positive video and a negative video, respectively. c) is a false-negative example.} For the illustration purpose, a threshold value of 0.5 is used to trigger the prediction of a future accident. In each example, sample frames of the video are shown on the top and the time series of the accident anticipation probability (i.e., the red colored curve) is displayed at the bottom. In a), the probability of anticipating an accident has reached the threshold at frame \#54, yielding 1.75 seconds TTA. From the sample frames, it is clear that the network successfully identifies the risk of accident when a few motorcycles are coming to the direct lane of the car involved in the accident later on. In b), the network does not predict an accident because the probability is always below 50\%. \textcolor{blue}{In c), the accident happens between a yellow car and a motorcycle, which are relatively far from the ego vehicle. The prediction score never exceeds the selected threshold value 0.5. The long-distance of the accident-involved agents from the ego vehicle might cause difficulty in extracting salient spatial-temporal relational dynamics of the two vehicles from their features. Therefore, the network missed predicting the accident ahead of time.}

\subsection{\textcolor{blue}{Performance of the Fusion Method}}
\label{sub:fusion_result}
The study evaluates the effectiveness of the proposed score-level late-fusion method on the DAD dataset. The top portion of TABLE \ref{tab:fusion} lists four candidate models for fusion, which are efficient and may complement one and another. The first two candidate models are learnt by the DSTA-network, \textcolor{blue}{which only differ in model parameters. DSTA(I) is the model yielding the best AP (see Fig. \ref{fig_mtta_ap}) and DSTA(II) is the one achieving the best P$_{80\text{R}}$}. The third candidate model is the best AP model of GCRNN \cite{bao2020uncertainty}, and its performance is attained by reproducing the model using the publicly available code\footnote{https://github.com/Cogito2012/UString}. The last candidate model is a modification of GCRNN, which is trained using the adaptive loss function of AdaLEA \cite{suzuki2018anticipating}. Here, AdaLEA provides another degree of flexibility to explore a higher AP. Compared to GCRNN, GCRNN+AdaLEA attains higher AP and P$_{80\text{R}}$, but shorter mTTA and TTA$_{80\text{R}}$. \textcolor{blue}{The last two candidate models have their respective network architecture, different than the DSTA-network.}

\begin{table*}[t]
\renewcommand{\arraystretch}{1.3}
    \caption{Assessment of the fusion method}
    \label{tab:fusion}
    \centering
    \begin{tabular}{c|l|l|r|r|r|r|r|r}
        \cline{2-9}
          \multicolumn{1}{c}{}&Model-1 & Model-2& $\bar{a}^{(1)}$ & $\bar{a}^{(2)}$ & AP (\%) & P$_{80\text{R}}$ (\%)  & mTTA (s) & TTA$_{80\text{R}}$ (s)   \\
         \hline
        \multirow{4}{*}{Individual}& DSTA (I)&-&-&-& 72.3 & 46.6 & 1.50 & 1.81 \\ 
         & DSTA (II)&-&-&-& 68.6 & 53.1 & 1.50 & 1.69 \\
         &GCRNN &-&-&-& 68.3 & 50.8 & 1.31 & 1.58 \\
         &GCRNN+AdaLEA&-&-&- & 69.4 & 50.6 & 1.02 & 1.18 \\
         \hline
         \multirow{3}{*}{Fusion}& DSTA (I)& GCRNN+AdaLEA & 0.72 & 0.55 & 74.1 & 53.1 & 1.08 & 1.18 \\
         & DSTA (I)& GCRNN & 0.40 & 0.20 & 73.8 & 48.9 & 1.47 & 1.60 \\
        & DSTA (I)& DSTA (II) & 0.42 & 0.50 & 72.6 & 56.4 & 1.50 & 1.70 \\
        \hline
    \end{tabular}
\end{table*}

The bottom portion of Table \ref{tab:fusion} summarizes three selected fusion results. Fusing the predicted scores of DSTA(I) and GCRNN+AdaLEA achieves a remarkably high AP, 74.1\%, with the corresponding mTTA 1.08 seconds. This result is achieved by selecting the prediction threshold 0.72 for DSTA(I), and 0.55 for GCRNN+AdaLEA. Compared to the first fusion result, the fusion of DSTA(I) with GCRNN slightly lowers the AP for 0.3\% but extends the mTTA for 0.39 seconds. Fusing DSTA(I) and DSTA(II) effectively leverages the $P_{80\text{R}}$ to 56.4\%, equivalent to a margin of 9.8\% compared to DSTA(I) and 3.3\% compared to DSTA(II). It is noted that the fusion of DSTA(I) and DSTA(II) provides a practical solution to real-world applications. This fusion result dominates all of the four individual models on the mTTA-AP diagram. Meantime, it provides probably the most attractive combination of precision and TTA, at 80\% recall. This fusion result further highlights the contribution of the proposed DSTA-network. \textcolor{blue}{The study found that the improvement from fusing more than two efficient models is marginal.}

\subsection{Impact of the Datasets}
TABLE \ref{tab:comparison} shows that performances of the accident anticipation models on the DAD dataset are significantly lower than on the CCD dataset. \textcolor{blue}{The literature has not revealed the underlying reasons for the observed dramatic difference. In attempt to reveal some of the unknowns, a thorough experimental study is conducted to evaluate the applicability and limitations of representative datasets for the early accident anticipation.} 

CCD is over 2.5 times larger than DAD. To determine if the performance gap is caused by the data size difference, \textcolor{blue}{a subset of CCD, which is in the same size as DAD and named as CCD-small, is created by randomly selecting videos from the original CCD dataset. Despite of its reduced size, the DSTA-network trained on CCD-small still achieves 98.51\% AP and 4.83 seconds mTTA. This verifies that the larger size of CCD is not the main reason for the superior performance of early accident anticipation on CCD.} 

\textcolor{blue}{DAD and CCD datasets are dramatically different. Accidents in the DAD dataset always start from the last 0.5 seconds of positive videos, only 10\% of the total video length. Most importantly, 90\% of the positive videos in DAD contains extremely complex urban traffic accidents, where ego-vehicles are not involved in accidents, and traffic agents involved in, or affected by accidents, may appear in the video for a very short time in relative to the full length of the training videos. In the CCD dataset, 53.4\% of the accidents are ego-vehicle involved. Changes in motion, size, and other appearance features in accident videos are relatively large.} The starting time of accidents in CCD is placed randomly in the last 2 seconds, and the mean starting time is in the last 1.28 seconds. These differences are important reasons that accident anticipation models have largely different performance on the two datasets. 
\begin{table}[htbp]
\renewcommand{\arraystretch}{1.3}
    \caption{The Impact of Dataset Differences}
    \label{tab:impact_dataset}
    \centering
    \begin{tabular}{l|c|r|r}
        \hline
         Training & Testing &AP(\%) & mTTA (s)  \\
         \hline
         DAD & CCD & 35.99 & 4.98 \\
         CCD & DAD & 35.74 & 4.99 \\
         CCD-small$\cup$DAD&CCD&{98.89}&{4.48}\\
         CCD-small$\cup$DAD&DAD&{69.67}&{1.40}\\
         \hline
    \end{tabular}
\end{table}
To determine if the above-discussed data differences would impact the generalization capability of the proposed DSTA-network, four experiments listed in TABLE \ref{tab:impact_dataset} are conducted. In these experiments, the frequency of DAD videos is reduced to 10 fps to be compatible with CCD. The first two experiments in this table are cross-testing. Both of the tests achieve a low AP value, near 36\%. The dramatically reduced AP implies a challenge for the early accident anticipation arising from differences in traffic scenes and accident types. The network is further trained on the dataset that mixes the CCD-small and the DAD training dataset. This model performs better than in the cross-testing experiments, achieving 98.89\% AP and 4.48 secconds mTTA when tested on the CCD dataset, and 69.67\% AP and 1.40 sec mTTA on DAD. But the performances are below those in TABLE \ref{tab:comparison}.

\textcolor{blue}{This experimental study verifies that the data foundation for the early accident anticipation is crucial. A comprehensive dataset is desired, which has balanced ego-vehicle involved and uninvolved accidents and diverse scene configurations from different regions and countries.}

\section{Conclusion}

This paper presented a novel end-to-end DSTA-network for the early anticipation of traffic accidents from widely deployed dashcams. Through designing a new model fusion method and analyzing existing datasets, this paper identifies opportunities for further advancing the early accident anticipation. 

\textcolor{blue}{Computer vision has been playing vital roles in traffic safety enhancement and autonomous driving. It can augment or substitute for human vision in the transportation system. Just like human vision that is one, but not the only, sensing and learning mechanism, computer-vision based early accident anticipation should be further integrated with other safety-enhancement technologies and methods such as LiDAR, internet of things, and transportation safety surrogate measurements (SSM) to deliver a multimodal, multifunctional system for the early accident anticipation. The convergence of these technologies with the proposed method is an important direction that will bring exciting opportunities for safety enhancement. This study has envisaged multiple lines of future work. The proposed DSTA-network has a common limitation as many other deep learning models, which is the unexplainability of the rationale behind the network's decision-making process. A follow-up study is to reveal the risk perception mechanism of the network to make it explainable and trustworthy to humans. In transportation, various surrogate safety measurements are effectively used to identify accident risks. An exciting future direction is to develop sensor-fusion based, SSM-informed deep learning networks for safety enhancement. This study also evidences the need for a strong data foundation for the accident anticipation.} Finally, accident anticipation is also a desired capability for mobile robots such as drones and remotely operated underwater vehicles. The application of the DSTA-network to those robots faces additional challenges. This paper lays a foundation for exploring the above-discussed exciting opportunities.

\section*{Acknowledgment}
Qin, Karim, and Li receive support from National Science Foundation (NSF) through the grant ECCS-\#2026357. Yin and Karim receive support from NSF through ECCS-\#2025929.

\ifCLASSOPTIONcaptionsoff
  \newpage
\fi



\bibliographystyle{IEEEtran}
\bibliography{ref}



\end{document}